\documentclass[runningheads]{llncs}

\usepackage[T1]{fontenc}
\usepackage{graphicx}
\usepackage{placeins}
\usepackage{amsmath,amssymb}
\usepackage{booktabs}
\usepackage{multirow}
\usepackage[numbers,sort]{natbib}
\usepackage{xcolor}
\usepackage{subcaption}
\usepackage{pgfplots}
\usepackage{float}
\pgfplotsset{compat=1.18}

\newcommand{\datasetname}{Genodisc}

\begin{document}

\title{Segmentation Pre-training for Label-Efficient \\ Lumbar Spine Degeneration Grading}

\titlerunning{Seg-Pretrain for Data-Efficient Spine Grading}

\author{%
Maria Monzon\inst{1,2}\and
Andrew Zisserman\inst{3} \and
Catherine R. Jutzeler\inst{1,2}\and
Amir Jamaludin\inst{3}
}
\authorrunning{M. Monzon et al.}

\institute{%
 Biomedical Data Science Lab, Dept. D-HEST, ETH Zurich, Zurich, Switzerland
\and Swiss Institute of Bioinformatics (SIB), Lausanne, 1015, Switzerland\\
\and
Visual Geometry Group, Dept.\ of Engineering Science,
University of Oxford, UK\\
\email{ mmonzon@ethz.ch}
}

\maketitle

\begin{abstract}
Automated assessment of degenerative pathology in the lumbar spine on magnetic resonance imaging (MRI) requires access to large-scale datasets of expert-annotated radiological gradings. 
In contrast, segmentation pseudo-labels can be generated by automated tools at negligible radiologist cost.
We examine whether pre-training on segmentation can effectively replace a fraction of the manual grading annotations required for downstream supervision.
We pre-train a 3D ResNet encoder to segment the vertebrae, intervertebral discs (IVDs), and the spinal canal, then fine-tune lightweight task-specific grading heads using different proportions of the available training data, ranging from $10\%$ to $100\%$.
On a multicentre dataset of ${\sim}2{,}000$ subjects across 11 pathologies, segmentation pre-training, achieving a Dice score of $0.94$ against pseudo-labels, improved the task-averaged (macro) one-vs-rest ROC-AUC at all  proportions.
With only 20\% of grading labels after pre-training, the method achieved near full-supervision performance, with the largest gains observed for either low-prevalence or spatially grounded pathologies.
\end{abstract}

\keywords{Segmentation Pre-training \and Label Efficiency \and Spine Grading  \and MRI}
\begin{figure}[!ht]
    \centering
    \includegraphics[width=0.92\linewidth]{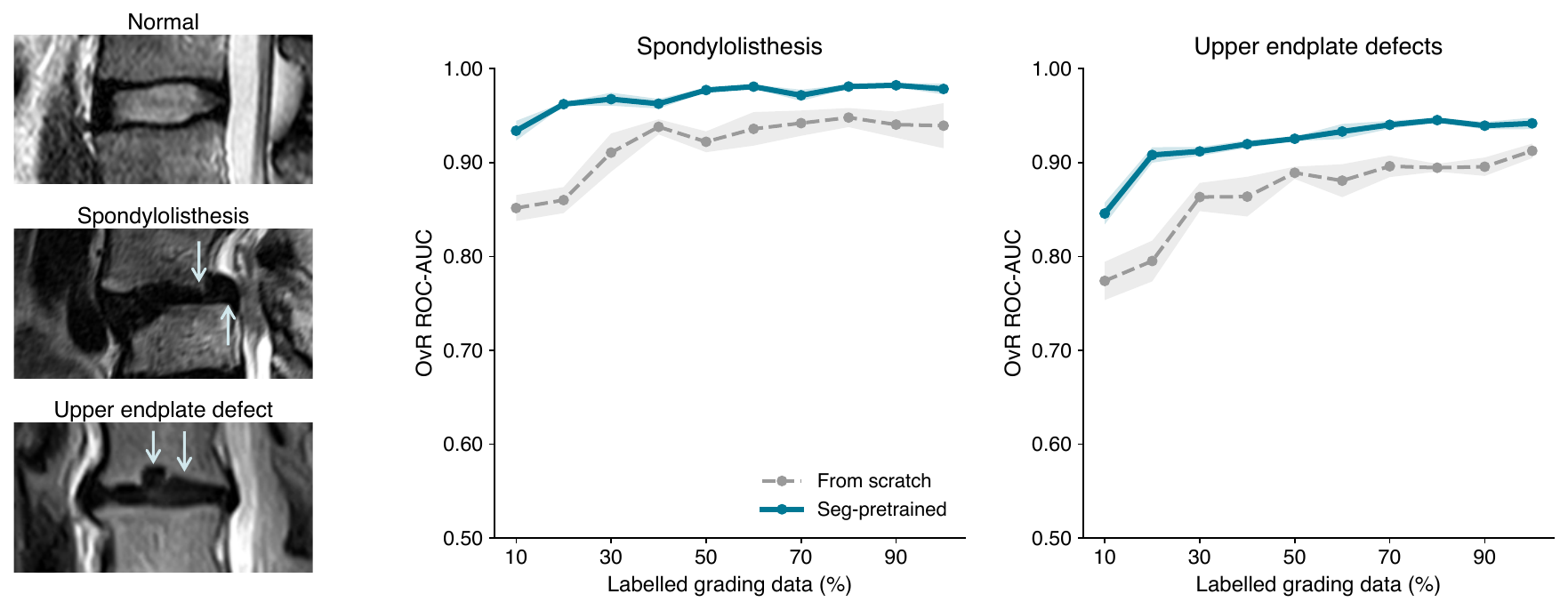}
    \caption{Example pathologies (left) and data-efficiency ROC-AUC curves (right) show near full-data performance using only 20\% of labels after pre-training. Shaded bands show the standard deviation over three random seeds.
}
    \label{fig:teaser}
\end{figure}

\section{Introduction}


Lumbar spine degeneration is a leading cause of low back pain, which affects
more than 800 million people globally~\cite{lbp-GBD2021}. Automated grading of MRI
has reached expert-level performance for several pathological conditions \cite{Jamaludin2017SpineNet,Windsor2024SpineNetv2},
including intervertebral disc (IVD) degeneration~\cite{Kowlagi2023-PfirmannBaseline}, central
canal stenosis~\cite{Lee2011-Stenosis}, and vertebral endplate
defects~\cite{Bassani2023-EndplateDefects}. 
However, the development of such models depends on large datasets of expert-provided ordinal annotations.
In lumbar MRI, degenerative labels are typically assigned independently for multiple pathologies at each IVD level, making dataset construction labor-intensive and difficult to scale.

In contrast, recent open-source methods are capable of producing robust segmentations of the vertebral bodies, IVD, and spinal canal, while demonstrating good generalisation across heterogeneous MRI acquisition protocols~\cite{Lu2018-DeepSpine,warszawer2025-TotalSpineSeg,moller2024-SpinePS}.
Crucially, the geometry these masks capture such as disc-height loss, lateral foraminal narrowing, and reduced spinal canal width, is precisely what drives many clinical radiological gradings of spinal degeneration.

In this work, we investigate whether pre-training on automatically derived
segmentation masks can substitute for a substantial portion of expert grading
annotations. On top of this segmentation pre-training we train lightweight per-task grading heads and quantify multi-pathology grading quality with imbalance-robust classification metrics.
By systematically varying the fraction of available grading labels from $10\%$ to
$100\%$, we characterise the annotation savings afforded by the segmentation: a pretrained 3D ResNet-18-UNet reaches the same macro one-vs-rest ROC-AUC with
substantially fewer manual annotations than a from-scratch model
(Fig.~\ref{fig:teaser}), and is more stable across seeds.
We also investigate predicting continuous gradings and observe the same label-efficiency pattern as with categorical classification.

\subsection{Related Work}

Dense voxel-wise supervision has emerged as a powerful pre-training paradigm in medical image analysis. 
By explicitly forcing the encoder to internalise structural boundaries and morphological geometry, supervised segmentation pre-training on large anatomical datasets yields transferable 3D representations that frequently outperform self-supervised alternatives on downstream tasks with scarce annotations~\cite{li2025-dl-transferlearning}. 
In the broader medical domain, auxiliary segmentation has successfully regularised encoders for cardiac diagnosis \cite{Snaauw2019-dl-transferlearning}.

In the context of the lumbar spine, recent automated frameworks such as DeepSPINE \cite{Lu2018-DeepSpine}, Bianque-Net \cite{Zheng2022-BianqueNet}, TotalSpineSeg~\cite{warszawer2025-TotalSpineSeg} and SpinePS~\cite{moller2024-SpinePS} reliably generate vertebral-body, intervertebral-disc, and/or spinal-canal masks across heterogeneous clinical cohorts at negligible cost. 
However, these segmentation outputs have traditionally been utilized merely to extract explicit geometric measurements for IVD degeneration~\cite{Neubert2013, Zheng2022-BianqueNet} and stenosis quantification~\cite{Natalia2020,Lu2018-DeepSpine}, or strictly as a spatial preprocessing step for localisation~\cite{Windsor2024SpineNetv2}, rather than as a learned inductive bias for spine degeneration learning.

Concurrently, automated grading of lumbar pathologies has been heavily explored via categorical classification~\cite{Jamaludin2017SpineNet,Windsor2024SpineNetv2}. 
A critical limitation uniting these approaches is their heavy reliance on large, exhaustively annotated datasets of expert ordinal labels. While transfer learning and self-supervised methods have sought to alleviate this bottleneck in general medical imaging \cite{li2025-dl-transferlearning, Hooper2023-dl-transferlearning}, the explicit use of automatically derived anatomical masks as a geometric pre-training strategy to reduce the clinical annotation budget remains systematically unexplored in lumbar spine degeneration.

\section{Segmentation-Supervised Spinal Grading}

Radiological grading of spinal pathologies is fundamentally based on spatial geometric biomarkers. 
Established clinical scales quantitatively evaluate vertebral morphology, IVD height, and spinal canal cross‑sectional area, all of which are spatially localised and structurally well defined~\cite{Pfirrmann2001-mri-discdeg,Lee2010-mri-foramstenosis,Koslosky2020-mri-spondyl}.
We therefore hypothesise that the latent representations required for accurate anatomical segmentation provide a sufficient basis for multi-task pathology grading.

The segmentation objective forces the encoder to learn structural boundaries and morphological deformations directly associated with spinal degeneration. 
Since anatomical masks are obtainable from automated tools at no radiologist cost, this provides a scalable geometric prior for grading.

\begin{figure}[tp]
    \centering
    \includegraphics[width=0.95\linewidth]{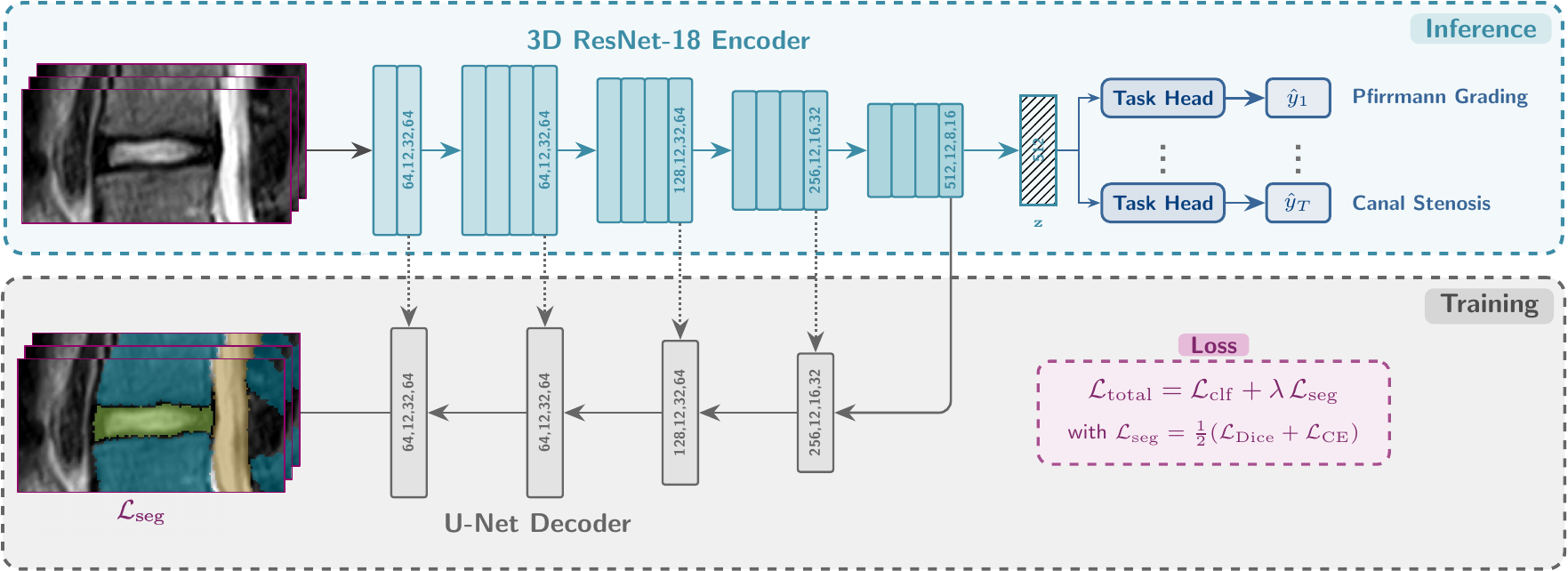}
    \caption{Segmentation-pretrained multi-task grading model. Segmentation is used during training, while grading is derived from encoder embeddings at inference.}
    \label{fig:architecture}
\end{figure}

We hypothesise that segmentation pre-training enhances ordinal grading, drastically reducing the need for manual labels. We propose a two-stage framework (Fig.~\ref{fig:architecture}) where an encoder is first pre-trained on vertebra, disc, and spinal canal masks, then fine-tuned on limited expert annotations using lightweight per-task grading heads. To evaluate this label efficiency, we compare our pretrained encoder against a randomly initialised baseline by systematically varying the proportion of available grading labels from $10\%$ to $100\%$.

\subsection{Segmentation Pre-training as a Spatial Supervision}
\label{sec:seg-pretrain}

We formulate segmentation pre-training as a means to distil 
clinically relevant spatial structure into a compact latent 
representation. Let $\mathbf{x}\in\mathbb{R}^{D\times H\times W}$ 
be a disc-centred IVD volume and $\mathbf{m}\in\{0,\ldots,C\}^{D\times H\times W}$ 
its voxel-wise anatomical mask with $C{=}3$ foreground classes 
(vertebral body, intervertebral disc, spinal canal). The encoder 
$f_\theta : \mathbb{R}^{D\times H\times W} \to \mathbb{R}^d$ is 
trained to map $\mathbf{x}$ to a latent embedding 
$\mathbf{z} = f_\theta(\mathbf{x})$ that captures the geometric 
structure necessary for precise anatomical delineation.
The encoder $f_\theta$ is jointly optimised with a U-Net decoder 
 under a composite Dice--CE loss:
\begin{equation}
    \mathcal{L}_{\text{seg}} = 
    \tfrac{1}{2}\bigl(\mathcal{L}_{\text{Dice}} + \mathcal{L}_{\text{CE}}\bigr),
    \label{eq:seg-loss}
\end{equation}
computed with foreground-macro Dice. 
The predicted masks also yield a
training-free read-out of mid-sagittal morphometry (disc height and Disc Height Index, wedging, canal antero-posterior width and listhesis offset) in millimetres via the voxel spacing, comparable across subjects.

\subsection{Multi-Task Grading Heads}
\label{sec:grading-heads}

After pre-training, the encoder $f_\theta$ is reused as the grading backbone. 
Each grading task $t$ has its own lightweight
\emph{classification} head $g_t:\mathbb{R}^{D}\!\to\!\mathbb{R}^{K_t}$ that maps the
shared embedding $\mathbf{z}=f_\theta(\mathbf{x})\in\mathbb{R}^{D}$ to $K_t$
ordinal-grade logits $\boldsymbol{\ell}_t = g_t(\mathbf{z})$, from which a softmax
yields the per-grade posterior.
The heads are trained with a class-weighted cross-entropy (CE). 
Fine-tuning is performed end-to-end on a label subset $\mathcal{D}_f \subset \mathcal{D}_{\text{grade}}$ under class-weighted cross-entropy, with an optional auxiliary segmentation loss gated by $\lambda_{\text{seg}}$:
\begin{equation}
    \mathcal{L} = \mathcal{L}_{\text{clf}} + \lambda_{\text{seg}}\,\mathcal{L}_{\text{seg}},
    \label{eq:grading-loss}
\end{equation}
where $\mathcal{L}_{\text{clf}}$ is the per-task class-weighted cross-entropy summed over the grading heads, $\mathcal{L}_{\text{seg}}$ the segmentation loss of Eq.~\ref{eq:seg-loss}, and $\lambda_{\text{seg}}$ the gate on the auxiliary segmentation term ($\lambda_{\text{seg}}{>}0$ keeps it active, $\lambda_{\text{seg}}{=}0$ switches it off).

\noindent\textbf{Continuous Severity Score.}
Spinal degeneration is inherently continuous, so discretising severity into ordinal
bins discards intra-grade nuance. Following recent work~\cite{monzon2026-spineranknet},
we extend the shared encoder with a continuous severity score: each task also
regresses its logits \(\boldsymbol{\ell}_t\) into a soft-bounded scalar
\(s_t \in [0,S]\). It is supervised by a pairwise ranking objective~\cite{Ahmed2021-DeepRankSVM,monzon2026-spineranknet}
combining a margin hinge over differently-graded pairs with a similarity regularisation term ($\lambda_{\text{sim}}{=}0.25$):

\begin{equation}
    \mathcal{L} = \lambda_{\text{seg}}\,\mathcal{L}_{\text{seg}} + \mathcal{L}_{\text{hinge}} + \lambda_{\text{sim}}\,\mathcal{L}_{\text{sim}}  ,
    \label{eq:rank-loss}
\end{equation}

%
The loss applies a grade-distance weighting that penalises larger-grade confusions
more heavily, optimised with hardest-pair sampling; full details are in the
supplementary material.




\section{Dataset and Implementation Details}
\label{sec:impl}

\noindent\textbf{Dataset.}
The \datasetname{} dataset~\cite{Jamaludin2017SpineNet} includes T1-weighted and T2-weighted sagittal lumbar MRIs from approximately 2,000 subjects acquired at five European centres with heterogeneous imaging parameters (0.2--3.0\,T, 0.25--0.9\,mm in-plane, 2.6--6.0\,mm slice spacing). Expert annotations cover $11$ ordinal grading tasks across six lumbar IVD levels (T12-L1 to L5-S1): disc degeneration (Pfirrmann grade, disc narrowing, herniation); central, left-foraminal, and right-foraminal stenosis; and vertebral abnormalities (spondylolisthesis, upper and lower endplate defects, anterior and posterior bulging). The dataset is strongly imbalanced, with normal grades accounting for 60--96\% of cases and severe grades 2--10\%. Patients are split 80/10/10 into train, validation, and test sets, stratified by centre and grade distribution.

\noindent \textbf{Preprocessing. }
Masks are obtained automatically using SpinePS~\cite{moller2024-SpinePS} and collapsed into four classes: background, vertebral body, intervertebral disc, and spinal canal. Disc-centred IVD regions are localised using SpineNetv2~\cite{Windsor2024SpineNetv2} and rotated and resampled with their masks to a shared geometry of $192\times320\times15$, which provides margin for augmentation. 
Final crops are extracted at $128\times256\times12$, using nearest-neighbour interpolation for masks to preserve discrete labels and avoid boundary artefacts. During training, 3D augmentations are applied: $\pm2$ slice shifts, flips ($p=0.5$), $\pm0.1$ intensity offsets, random in-plane translations of up to $32/24$ px, $\pm10\%$ scaling, and $\pm15^\circ$ in-plane rotation.

\noindent\textbf{Architecture.}
The encoder $f_\theta$ is a 3D ResNet with asymmetric strides $(1,s,s)$ that subsample only the in-plane axes and preserve slice depth. Global average pooling yields a shared embedding $\mathbf{z}\in\mathbb{R}^D$, with $D=512$ for ResNet-18/34 and $D=2048$ for ResNet-50. During segmentation pre-training, a symmetric U-Net decoder is attached to produce a dense 4-class voxel map via bilinear upsampling and encoder skip connections. The shared embedding feeds 14 task-specific classification heads, each a two-layer MLP $\mathbb{R}^D\!\to\!\mathbb{R}^{256}\!\to\!\mathbb{R}^{K_t}$ with Linear--ReLU--Dropout--Linear. Of these, the 11 reported ordinal grading tasks additionally use ranking heads $\varphi_t:\mathbb{R}^{K_t}\!\to\!\mathbb{R}^{8}\!\to\!\mathbb{R}$; the remaining three heads---disc level and upper/lower Modic changes---are not evaluated as pathologies but retained as auxiliary tasks that regularise the shared encoder in both the classification and ranking settings.

\noindent\textbf{Training.}
All phases use AdamW (weight decay~$10^{-5}$), a cosine schedule and batch
size~$32$. The encoder and a symmetric U-Net decoder are first pretrained for up
to 200~epochs (lr~$10^{-3}$) to reproduce the SpinePS masks under the
Dice--cross-entropy loss of Eq.~\eqref{eq:seg-loss}. The grading model is then
fine-tuned in two stages: a 30-epoch frozen-encoder head warm-up (lr~$10^{-3}$)
then 100~epochs end-to-end (lr~$10^{-4}$) under class-weighted cross-entropy,
keeping the lowest-validation-loss checkpoint; the from-scratch baseline runs
end-to-end only for a matched 130-epoch budget (lr~$10^{-3}$).


\noindent \textbf{Evaluation.}
To quantify the annotation savings we fine-tune the grader at label fractions from
$10\%$ to $100\%$, comparing the segmentation-pretrained encoder (Seg) against an
identically trained random initialisation (Scratch) over three seeds (7/42/123).
On the held-out test set we report classification metrics macro-averaged over the
$11$ grading tasks: macro one-vs-rest ROC-AUC, Matthews correlation coefficient
(MCC) and balanced accuracy, prevalence-robust under the strong class imbalance.
For the continuous-ranking variant, we additionally report Quadratic Weighted Kappa (QWK), and macro cumulative ROC-AUC of the severity score.
\section{Experimental Results}
\label{sec:results}

\subsection{Segmentation Pre-training}

\begin{figure}[tb]
    \centering
    \includegraphics[width=\linewidth]{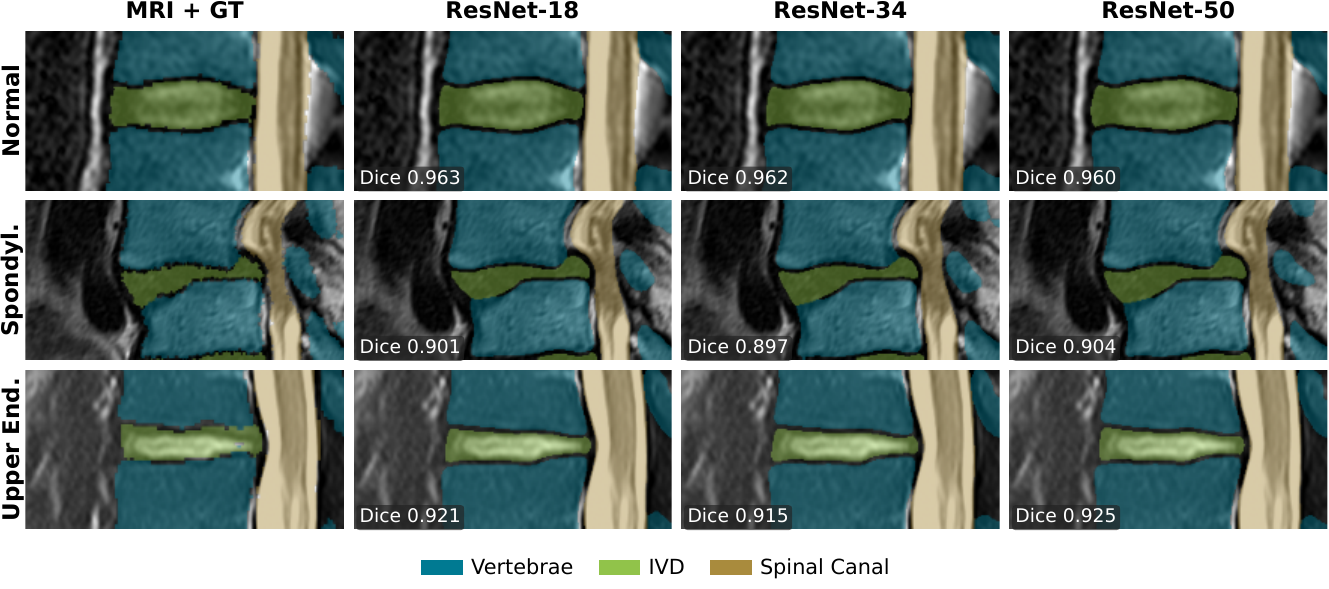}
\caption{Segmentation across backbones, spondylolisthesis and upper-endplate defect examples, with ground-truth overlay, predictions, and test mean Dice.}
\label{fig:seg-by-backbone}
\end{figure}
Table~\ref{tab:results-dice} shows near-identical segmentation performance across all three backbones, with mean Dice score of $0.94\pm0.03$ on the test set, computed against automatically generated SpinePS masks.
Qualitative results in Fig.~\ref{fig:seg-by-backbone} indicate that the predicted segmentation masks exhibit comparable accuracy across cases and severity levels.
We therefore use the lightweight ResNet-18 in the remaining experiments to minimise inference cost.

\begin{table}[tb]
    \caption{Test-set segmentation Dice (\%, mean\,$\pm$\,std) by backbone and structure.}
    \centering
    \setlength{\tabcolsep}{10pt}
    \begin{tabular}{l|ccc|c}
        \toprule
        Backbone & Vertebrae & IVD & Spinal canal & Mean \\
        \midrule
        ResNet-18-UNet  & 95.20\,$\pm$\,2.06 & 91.75\,$\pm$\,7.51 & 94.91\,$\pm$\,2.99 & 93.95\,$\pm$\,3.44 \\
        ResNet-34-UNet  & 95.14\,$\pm$\,2.06 & 91.65\,$\pm$\,7.63 & 94.90\,$\pm$\,2.90 & 93.90\,$\pm$\,3.45 \\
        ResNet-50-UNet  & 95.18\,$\pm$\,2.21 & 91.72\,$\pm$\,7.62 & 94.98\,$\pm$\,2.94 & 93.96\,$\pm$\,3.53 \\
        \bottomrule
    \end{tabular}
    \label{tab:results-dice}
\end{table}

\subsection{Data-Efficient Ordinal-Grade Discrimination}

\begin{figure}[t!]
    \centering
    \includegraphics[width=\linewidth]{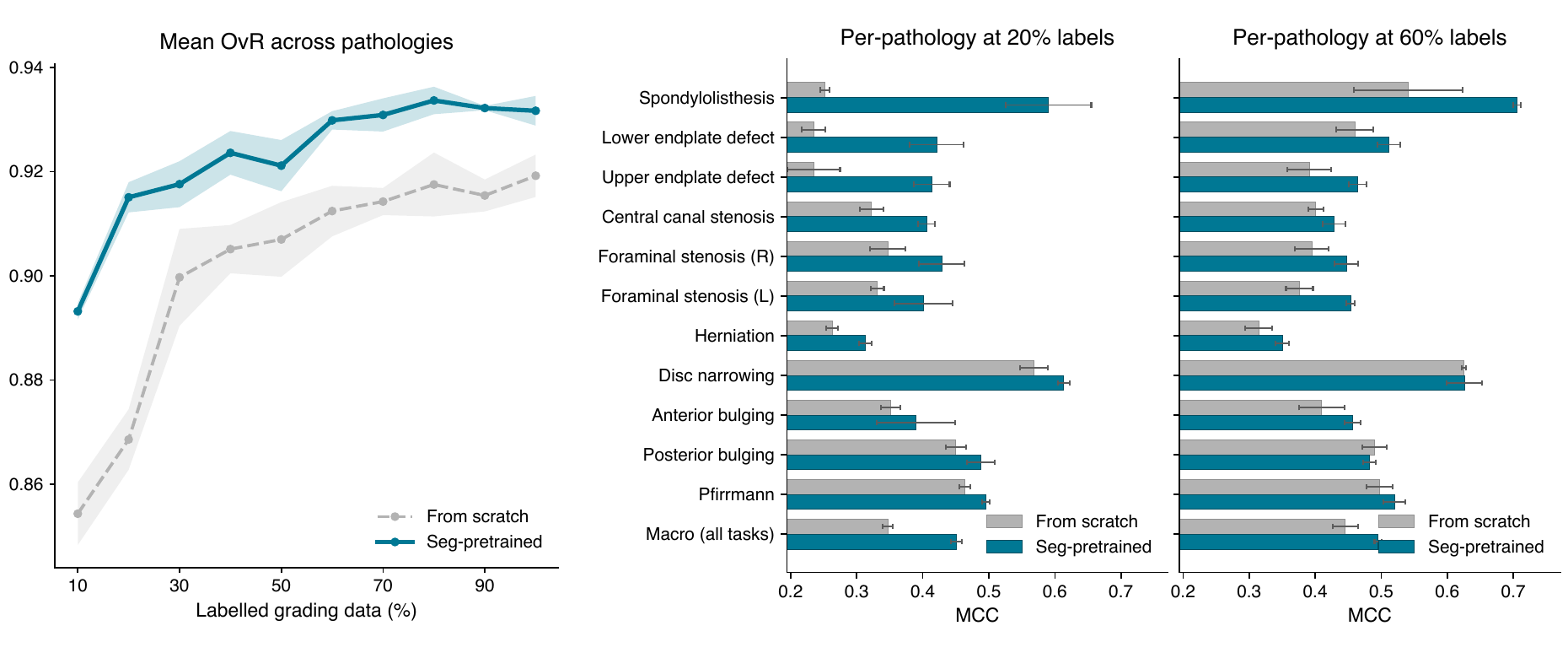}
  \caption{\textbf{Label efficiency of segmentation pre-training comparison.} (left) Macro one-vs-rest ROC-AUC vs.\ labeled-data fraction (mean $\pm$ s.d., $n{=}3$ seeds), pretrained (petrol) vs.\ scratch (grey). (middle, right) Per-pathology Matthews correlation coefficient
(MCC) at $20\%$ and $60\%$ label supervision.
} \label{fig:dataeff-overall}
\end{figure}

Segmentation-based pre-training improves the label-efficiency trade-off.
Averaged over all eleven ordinal tasks, the segmentation-pre-trained 3D ResNet-18 consistently outperforms the from-scratch baseline in terms of macro one-vs-rest ROC-AUC at all fractions of grading labels (Fig.~\ref{fig:dataeff-overall}, left), even improving from $0.919$ to $0.932$ under full supervision. 

\begin{table}[b!]
\centering
\caption{Per-pathology one-vs-rest ROC-AUC for from-scratch and segmentation pre-trained models at 20\%, 50\%, and 100\% labels. Entries are mean ROC-AUC with standard deviation ($10^{-2}$ units) over $n{=}3$ seeds.}
\label{tab:perpath-auc}
\scriptsize
\setlength{\tabcolsep}{4pt}
\resizebox{\columnwidth}{!}{%
\begin{tabular}{lcccccc}
\toprule
 & \multicolumn{2}{c}{\textbf{20\%}} & \multicolumn{2}{c}{\textbf{50\%}} & \multicolumn{2}{c}{\textbf{100\%}} \\
\cmidrule(lr){2-3}\cmidrule(lr){4-5}\cmidrule(lr){6-7}
\textbf{Pathology} & Scratch & Pre-train & Scratch & Pre-train & Scratch & Pre-train \\
\midrule
Spondylolisth. & $0.860{\scriptscriptstyle\,\pm1.4}$ & $0.962{\scriptscriptstyle\,\pm0.2}$ & $0.922{\scriptscriptstyle\,\pm1.1}$ & $0.977{\scriptscriptstyle\,\pm0.3}$ & $0.939{\scriptscriptstyle\,\pm2.4}$ & $0.978{\scriptscriptstyle\,\pm0.6}$ \\
Endplate defect (L) & $0.843{\scriptscriptstyle\,\pm0.6}$ & $0.929{\scriptscriptstyle\,\pm1.5}$ & $0.932{\scriptscriptstyle\,\pm1.0}$ & $0.940{\scriptscriptstyle\,\pm0.5}$ & $0.944{\scriptscriptstyle\,\pm0.2}$ & $0.954{\scriptscriptstyle\,\pm0.2}$ \\
Endplate defect (U) & $0.795{\scriptscriptstyle\,\pm2.2}$ & $0.908{\scriptscriptstyle\,\pm0.8}$ & $0.889{\scriptscriptstyle\,\pm0.7}$ & $0.925{\scriptscriptstyle\,\pm0.2}$ & $0.912{\scriptscriptstyle\,\pm0.8}$ & $0.942{\scriptscriptstyle\,\pm0.6}$ \\
Canal stenosis & $0.929{\scriptscriptstyle\,\pm0.9}$ & $0.957{\scriptscriptstyle\,\pm0.0}$ & $0.955{\scriptscriptstyle\,\pm0.5}$ & $0.965{\scriptscriptstyle\,\pm0.5}$ & $0.958{\scriptscriptstyle\,\pm0.8}$ & $0.967{\scriptscriptstyle\,\pm0.4}$ \\
Foraminal sten. (R) & $0.900{\scriptscriptstyle\,\pm1.1}$ & $0.935{\scriptscriptstyle\,\pm0.7}$ & $0.920{\scriptscriptstyle\,\pm0.8}$ & $0.925{\scriptscriptstyle\,\pm1.2}$ & $0.923{\scriptscriptstyle\,\pm0.4}$ & $0.941{\scriptscriptstyle\,\pm0.4}$ \\
Foraminal sten. (L) & $0.887{\scriptscriptstyle\,\pm0.4}$ & $0.924{\scriptscriptstyle\,\pm0.9}$ & $0.911{\scriptscriptstyle\,\pm1.1}$ & $0.921{\scriptscriptstyle\,\pm1.8}$ & $0.919{\scriptscriptstyle\,\pm0.8}$ & $0.936{\scriptscriptstyle\,\pm0.3}$ \\
Herniation & $0.845{\scriptscriptstyle\,\pm0.5}$ & $0.878{\scriptscriptstyle\,\pm0.4}$ & $0.862{\scriptscriptstyle\,\pm1.0}$ & $0.870{\scriptscriptstyle\,\pm1.5}$ & $0.891{\scriptscriptstyle\,\pm0.5}$ & $0.890{\scriptscriptstyle\,\pm1.9}$ \\
Narrowing & $0.904{\scriptscriptstyle\,\pm0.6}$ & $0.924{\scriptscriptstyle\,\pm0.3}$ & $0.924{\scriptscriptstyle\,\pm0.4}$ & $0.932{\scriptscriptstyle\,\pm0.2}$ & $0.932{\scriptscriptstyle\,\pm0.1}$ & $0.932{\scriptscriptstyle\,\pm0.5}$ \\
Bulging (Ant) & $0.838{\scriptscriptstyle\,\pm0.9}$ & $0.863{\scriptscriptstyle\,\pm2.2}$ & $0.869{\scriptscriptstyle\,\pm0.8}$ & $0.893{\scriptscriptstyle\,\pm0.3}$ & $0.883{\scriptscriptstyle\,\pm1.1}$ & $0.902{\scriptscriptstyle\,\pm0.5}$ \\
Bulging (Post) & $0.878{\scriptscriptstyle\,\pm0.9}$ & $0.899{\scriptscriptstyle\,\pm0.6}$ & $0.897{\scriptscriptstyle\,\pm1.1}$ & $0.896{\scriptscriptstyle\,\pm0.7}$ & $0.908{\scriptscriptstyle\,\pm0.4}$ & $0.905{\scriptscriptstyle\,\pm0.4}$ \\
Pfirrmann & $0.876{\scriptscriptstyle\,\pm0.4}$ & $0.888{\scriptscriptstyle\,\pm0.3}$ & $0.895{\scriptscriptstyle\,\pm0.2}$ & $0.887{\scriptscriptstyle\,\pm0.5}$ & $0.901{\scriptscriptstyle\,\pm0.5}$ & $0.901{\scriptscriptstyle\,\pm0.5}$ \\
\midrule
\textbf{Macro (11 tasks)} & $0.869{\scriptscriptstyle\,\pm0.9}$ & $0.915{\scriptscriptstyle\,\pm0.7}$ & $0.907{\scriptscriptstyle\,\pm0.8}$ & $0.921{\scriptscriptstyle\,\pm0.7}$ & $0.919{\scriptscriptstyle\,\pm0.7}$ & $0.932{\scriptscriptstyle\,\pm0.6}$ \\
\bottomrule
\end{tabular}}
\end{table}

Per-pathology detailed results (Table~\ref{tab:perpath-auc}) show an
advantage of segmentation pre-training over training from scratch for all eleven tasks with $20\%$ label supervision
and for most tasks at $50\%$, with the largest improvements concentrated in the rare, geometry-defined tasks whereas the gains were smaller for common pathologies for which both models already performed well. The only exceptions are Pfirrmann at $50\%$, and herniation and posterior bulging at $100\%$ (Table~\ref{tab:perpath-auc}).

This pattern is especially clear for
spondylolisthesis ($0.860 \rightarrow 0.962$ ROC-AUC at $20\%$ supervision) and the upper-
and lower-endplate defects ($0.795 \rightarrow 0.908$ and
$0.843 \rightarrow 0.929$), indicating that the pre-trained encoder is most
beneficial when the target depends strongly on localised anatomy (Fig.~\ref{fig:teaser}).
The same effect is even clearer in MCC (Fig.~\ref{fig:dataeff-overall}), where separation is strongest for these minority-class tasks at $20\%$ labels and persists at $60\%$.

\subsubsection{Continuous Ranking of Severity.}

\begin{figure}[t!]
    \centering
    \includegraphics[width=\linewidth]{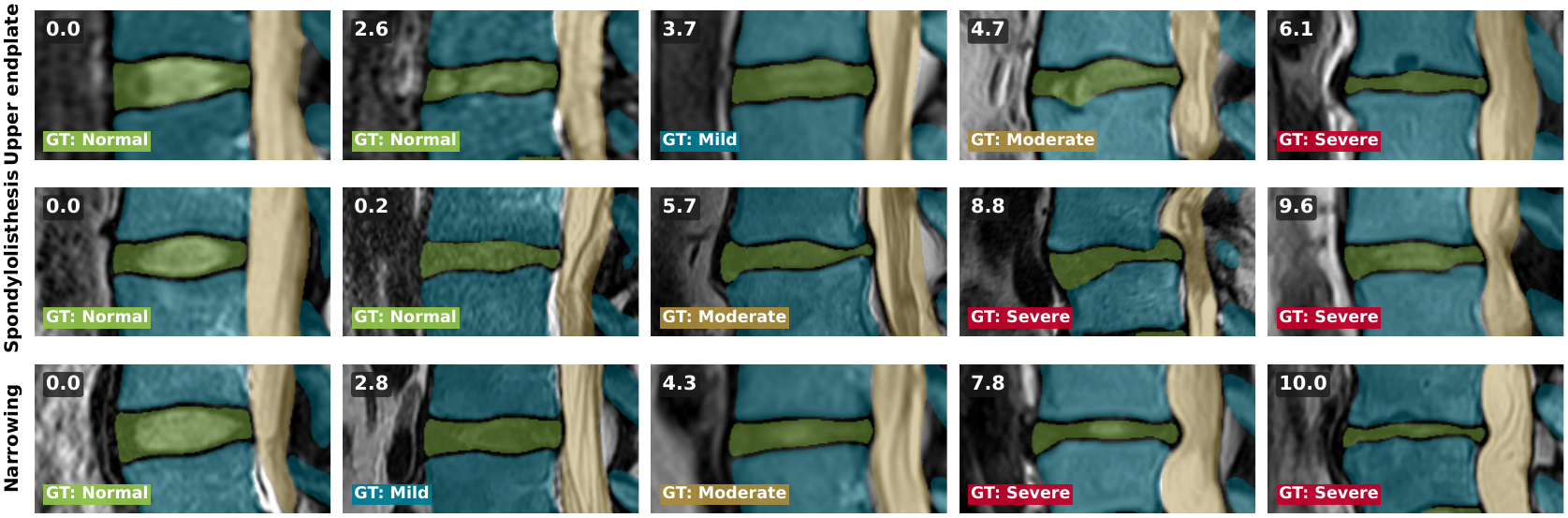}
\caption{\textbf{Continuous severity ranking with predicted segmentations.} Test IVDs ordered by predicted severity score (top-left; ground-truth bottom-left). The score increases monotonically, correctly ordering severity even when adjacent clinical grades overlap. Disc narrowing examples are all at the L4--L5 level.}
    \label{fig:ranking-qualitative}
\end{figure}

Segmentation pre-training also improves a continuous per-task severity-ranking
head in place of the categorical one, and the prior again helps most when labels
are scarce: at $10\%$ labels, it lifts the macro
quadratic-weighted $\kappa$ from $0.55$ to $0.68$ and the cumulative ROC-AUC from
$0.92$ to $0.95$, remaining ahead at $20\%$ (QWK $0.69$ vs.\ $0.63$).
Qualitatively, Fig.~\ref{fig:ranking-qualitative} shows a monotonic increase in
predicted severity score with grade for upper-endplate defects,
spondylolisthesis, and disc narrowing.

\section{Discussion and Conclusion}

This work shows that segmentation pre-training provides an effective anatomical supervision for data-efficient multi-task spine grading. 
Segmentation pre-training, learned from automatically generated masks with similar Dice scores across the evaluated backbones (0.94 against SpinePS-generated pseudo-labels), transfers effectively to degeneration grading.
By leveraging this spatial prior, a seg-pretrained 3D ResNet-18 reaches the macro ROC-AUC of a fully supervised from-scratch baseline using only 20\% of the labels.
The gain is most pronounced for rare, spatially grounded pathologies such as spondylolisthesis and endplate defects, suggesting that the benefit arises from clinically relevant geometric structure rather than generic regularization.

These findings suggest that segmentation pre-training is particularly well suited to grading tasks driven by localised morphology. 
However, it is limited by evaluation on a single cohort, despite being multicentre, and automatically generated pre-training masks.
Future work should assess external validation and whether the same strategy transfers across cohorts, MRI acquisition parameters, and other anatomically grounded degeneration tasks, such as knee osteoarthritis.

\subsubsection*{Acknowledgements.}
This project was supported by a grant \#380 (Jutzeler, Manjaly) from the
Strategic Focus Area ``Personalized Health and Related Technologies (PHRT)''
of the ETH Domain (Swiss Federal Institute of Technology). This project was also funded by the EPSRC programme grant Visual AI (EP/T025872/1). We thank our clinical collaborators Prof.\ J.\ Fairbank, Dr.\ J.\ Urban, Dr.\ S.\ Ather, and Prof.\ I.\ McCall.
AI-assisted tools (Claude, Perplexity, Writefull) were used for coding and grammar refinement; the authors reviewed all output and take full responsibility for the content.

\subsubsection*{Disclosure of Interests.}
C.R.J. is a scientific consultant to AbbVie. This role is unrelated to, and did not influence, the design, analysis, or reporting of the present study.
The remaining authors have no competing interests to declare.

\renewcommand{\bibsection}{\section*{\refname}}
\bibliographystyle{splncs04}
\bibliography{references}

\clearpage
\appendix

\begin{center}
  {\large\bfseries Supplementary Material}
\end{center}

\suppressfloats[t]

\makeatletter
\@addtoreset{figure}{section}
\@addtoreset{table}{section}
\makeatother
\renewcommand{\thefigure}{\thesection\arabic{figure}}
\renewcommand{\thetable}{\thesection\arabic{table}}
\section{Dataset}
\label{app:dataset}
The \datasetname{} grading distribution is summarized in
Table~\ref{tab:label_distribution}. Pfirrmann, the only five-grade task,
comprises Grade~I: 3911 (32.3\%), Grade~II: 1763 (14.6\%), Grade~III:
2824 (23.4\%), Grade~IV: 2424 (20.0\%), and Grade~V: 1169 (9.7\%).
\begin{table}[h!]
\centering
\caption{Pathology grade frequency per grading task for the complete \datasetname{} dataset (all splits).}
\label{tab:label_distribution}
\small
\setlength{\tabcolsep}{4pt}
\begin{tabular}{lrrrr}
\toprule
Task & Normal (\%)& Mild (\%)& Moderate (\%)& Severe (\%)\\
\midrule
Narrowing             & 7246 (60.0) & 1387 (11.5) & 2198 (18.2) & 1242 (10.3) \\
Canal stenosis        & 11364 (94.1) & 292 (2.4) & 186 (1.5) & 236 (2.0) \\
Upper Endplate defect   & 11045 (91.4) & 950 (7.9) & 61 (0.5) & 26 (0.2) \\
Lower Endplate defect  & 11023 (91.3) & 921 (7.6) & 88 (0.7) & 47 (0.4) \\
Foraminal stenosis Left   & 10867 (89.9) & 676 (5.6) & 275 (2.3) & 265 (2.2) \\
Foraminal stenosis Right  & 10841 (89.7) & 753 (6.2) & 244 (2.0) & 243 (2.0) \\
Herniation            & 10758 (89.0) & 639 (5.3) & 424 (3.5) & 262 (2.2) \\
Bulging (Ant.)        & 8953 (74.1) & 2050 (17.0) & 765 (6.3) & 315 (2.6) \\
Bulging (Post.)       & 8451 (69.9) & 2607 (21.6) & 828 (6.9) & 198 (1.6) \\
Spondylolisthesis     & 11609 (96.2) & 434 (3.6) & 30 (0.2) & 7 (0.1) \\
\bottomrule
\end{tabular}
\end{table}
\section{Auxiliary Segmentation Loss and Continuous Ranking}
\label{app:ablations}\label{app:ranking}\label{sec:supp-ranking}

\noindent\textbf{Auxiliary loss.}
Table~\ref{tab:aux-ablation} summarizes the effect of keeping the
segmentation loss active ($\lambda_{\text{seg}}{=}0.5$) as an auxiliary term during fine-tuning.
Gains are small but most pronounced
in regimes with low percentages of grading supervision.

\begin{table}[t]
\centering
\caption{Mean one-vs-rest ROC-AUC by auxiliary segmentation-loss weight $\lambda_{\text{seg}}$.}
\label{tab:aux-ablation}
\footnotesize
\setlength{\tabcolsep}{4pt}
\begin{tabular}{c ccc ccc}
\toprule
& \multicolumn{3}{c}{Scratch} & \multicolumn{3}{c}{Seg-pretrained} \\
\cmidrule(lr){2-4}\cmidrule(lr){5-7}
$\lambda_{\text{seg}}$ & 20\% & 50\% & 100\% & 20\% & 50\% & 100\% \\
\midrule
0   & 0.875 & 0.916 & 0.917 & 0.911 & 0.924 & 0.928 \\
0.5 & 0.895 & 0.919 & 0.924 & 0.924 & 0.935 & 0.942 \\
\bottomrule
\end{tabular}
\end{table}

\noindent\textbf{Continuous ranking.} The
embedding $z$ also yields a continuous severity score $s_t\in[0,S]$ read from the task logits, supervised by a severity-weighted
hinge~\cite{Ahmed2021-DeepRankSVM,monzon2026-spineranknet} over differently-graded pairs $\mathcal{P}_t$,
\begin{equation}
    \mathcal{L}_{\text{hinge}}^t =
    \frac{1}{|\mathcal{P}_t|}\!\!\sum_{(\mathbf{x}_i,\mathbf{x}_j)\in\mathcal{P}_t}\!\!
    w_{ij}^t\,\max\bigl(0,\, m_{ij}^t - r_{ij}^t(s_{ti} - s_{tj})\bigr),
    \label{eq:hinge}
\end{equation}
with $r_{ij}^t\in\{-1,0,1\}$, weight $w_{ij}^t \propto |y_{ti}-y_{tj}|/(K_t{-}1)$ and
margin $m_{ij}^t = m_0 + m_1|y_{ti}-y_{tj}|/(K_t{-}1)$ ($m_0{=}2.0$, $m_1{=}1.5$), plus
a similarity term
$\mathcal{L}_{\text{sim}}^t = \tfrac{1}{N_t}\textstyle\sum_{i}(s_{ti} - y_{ti}S/(K_t{-}1))^2$
that ties each score to $y_{ti}S/(K_t{-}1)$, giving within-grade compactness and
inter-grade calibration. We refer to~\cite{monzon2026-spineranknet} for the full derivation of the continuous score.
With
$\mathcal{L}_{\text{rank}}=\sum_t(\mathcal{L}_{\text{hinge}}^t+\lambda\mathcal{L}_{\text{sim}}^t)$,
$\lambda{=}0.25$, segmentation pre-training leads on cumulative ROC-AUC, QWK
(Table~\ref{tab:macro-efficiency}).

\begin{table}[t]
\centering
\caption{Macro cumulative ROC-AUC and QWK from the continuous severity score.}
\label{tab:macro-efficiency}
\footnotesize
\setlength{\tabcolsep}{4pt}
\begin{tabular}{c ccc ccc}
\toprule
& \multicolumn{3}{c}{Scratch} & \multicolumn{3}{c}{Seg-pretrained} \\
\cmidrule(lr){2-4}\cmidrule(lr){5-7}
Metric & 20\% & 50\% & 100\% & 20\% & 50\% & 100\% \\
\midrule
AUC & 0.941 & 0.953 & 0.959 & \textbf{0.956} & \textbf{0.960} & \textbf{0.961} \\
QWK & 0.629 & 0.671 & 0.691 & \textbf{0.688} & \textbf{0.708} & \textbf{0.717} \\
\bottomrule
\end{tabular}
\end{table}

\section{Morphological Parameter Extraction}
\label{app:morphology}
\begin{figure}[b!]
    \centering
    \includegraphics[width=0.74\linewidth]{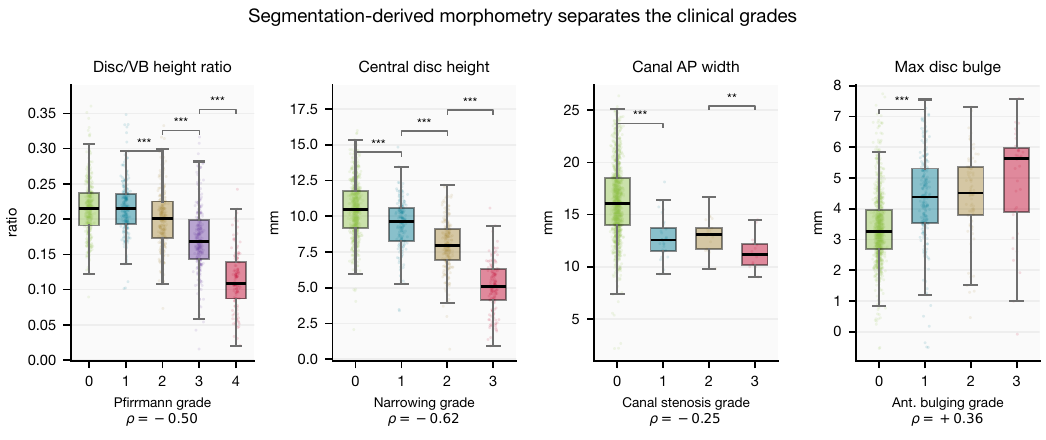}
    \caption{Segmentation-derived geometrical parameters by ground-truth grade, extracted on the test set.
    Panels show the four descriptors with the strongest grading
    associations. Axes report Spearman~$\rho$. Significant adjacent-grade
    differences (Mann--Whitney $U$) corrected (Benjamini--Hochberg FDR) are marked ($^{*}q<0.05$, $^{**}q<0.01$, $^{***}q<0.001$).}
    \label{fig:morph-box}
\end{figure}

From the predicted masks we extract interpretable geometric descriptors (mm) by fitting each structure's principal axes: disc heights perpendicular to the major axis, canal AP width along the minor axis, and maximum disc bulge as the largest radial distance the boundary extends past a fitted ellipse.
Associations with grade use Spearman's $\rho$ (Fig.~\ref{fig:morph-box}).
As descriptors are non-normal (Shapiro--Wilk $p<0.05$), grade separation is assessed non-parametrically: a Kruskal--Wallis test per parameter, then adjacent-grade Mann--Whitney $U$ tests, both FDR-corrected (Benjamini--Hochberg).

\end{document}